%% file: naacl2021.tex
\pdfoutput=1

\documentclass[11pt]{article}

\usepackage[]{naacl2021}

\usepackage{times}
\usepackage{latexsym}

\usepackage[T1]{fontenc}

\usepackage[utf8]{inputenc}

\usepackage{microtype}
\usepackage{amsfonts}
\usepackage{amsmath}
\usepackage{graphicx}
\usepackage{multirow}
\usepackage{caption}
\usepackage{subfigure}
\usepackage{enumitem}
\usepackage{float}
\usepackage{makecell}
\usepackage{placeins}
\usepackage{amssymb}
\usepackage{booktabs}
\usepackage{tabu}

\usepackage{hyperref}

\title{Mitigating Temporal-Drift: A Simple Approach to Keep NER Models Crisp}

\author{
Shuguang Chen \textsuperscript{$\dagger$}, 
Leonardo Neves \textsuperscript{$\ddagger$} \and 
Thamar Solorio \textsuperscript{$\dagger$} \\
University of Houston \textsuperscript{$\dagger$} 
\\ Snap Research \textsuperscript{$\ddagger$}\\
\texttt{schen52@uh.edu}, \texttt{lneves@snap.com} \and \texttt{tsolorio@uh.edu}}

\begin{document}
\maketitle
\begin{abstract}
Performance of neural models for named entity recognition degrades over time, becoming stale. This degradation is due to temporal drift, the change in our target variables' statistical properties over time. This issue is especially problematic for social media data, where topics change rapidly. In order to mitigate the problem, data annotation and retraining of models is common. Despite its usefulness, this process is expensive and time-consuming, which motivates new research on efficient model updating. In this paper, we propose an intuitive approach to measure the potential trendiness of tweets and use this metric to select the most informative instances to use for training. We conduct experiments on three state-of-the-art models on the Temporal Twitter Dataset. Our approach shows larger increases in prediction accuracy with less training data than the alternatives, making it an attractive, practical solution. \footnote{We release the code at \url{https://github.com/RiTUAL-UH/trending_NER}.}
\end{abstract}

\section{Introduction}
Prediction performances of live machine learning systems degrade over time due to changes in the statistical properties of the data used for training them. This degradation, also known as temporal drift, happens in different ML tasks, including named entity recognition (NER). Due to the nature of the task, authors also call this language drift \citep{fromreide-etal-2014-crowdsourcing, derczynski-etal-2015-usfd}.
Temporal drift effects are amplified in social media. Due to the ecosystem's very nature, topics reflect events and interests of a diverse user base and are continuously and rapidly evolving. To study the impact of language drift, we focus our analysis on the case of NER on Twitter data.
Emerging and Trending topics are an essential part of Twitter. They
change quite rapidly, reflecting diverse topics and world events of interest. Entities are a significant component of these changes, generating a diverse set of NE tokens. These ever-evolving topics pose a challenge as new entities frequently arise. The new entities are especially problematic as they might not exist in our previous vocabulary or can potentially transform the meaning of a previously observed term. Figure \ref{fig:example_of_tweets} shows tweets that include the emerging topic 'US'. After the release of the film, the topic 'US' became trending and aroused wide discussion. 
To mitigate the impact of temporal drift, we investigate how to effectively and efficiently adapt an already trained NER model to sustain prediction performance over time. We propose an intuitive approach to measure the trendiness of tweets and use this metric to select the most informative instances for retraining. We show that labeling instances based on this approach can yield better downstream performance than randomly sampling tweets for annotation.
\input{figures/example_of_tweets}

Note that topics such as semantic shift \citep{hamilton-etal-2016-cultural, rosenfeld-erk-2018-deep} and active learning \citep{Sinha2019VariationalAA, Kirsch2019BatchBALDEA} are related to the work we present here. In semantic shift, the core problem is how to trace temporal changes in lexical semantics, including linguistic drifts and cultural shifts. Unlike this task, our goal is to leverage the emergence of trends to guide an already trained model.

In active learning, researchers have focused on incremental annotation of instances by selecting the most informative ones. The goal is to achieve better results than random sampling. Multiple approaches exist to measure the informativeness of data points, but all of them are domain agnostic \citep{Sinha2019VariationalAA, Kirsch2019BatchBALDEA}. Our proposed solution is more straightforward than using uncertainty in ensembles or adversarial networks. However, it effectively increases model performance, and, similar to active learning approaches, it is more efficient than random sampling.

To summarize, we make the following contributions:
\begin{enumerate}[topsep=0pt,itemsep=-1ex,partopsep=1ex,parsep=1ex]
\item We propose an approach to measure the potential trendiness of tweets for selecting the most informative training samples. 

\item We conduct extensive experiments and demonstrate the effectiveness of our approach for retraining a NER model.
\end{enumerate}{}

\section{Emerging Trend Detection}
We want to exploit social media's inherent characteristics \citep{Benhardus2013StreamingTD, Mathioudakis2010TwitterMonitorTD}, with a focus on Twitter, to update model parameters efficiently. We assume that named entities associated with posts that are likely to become trends will be more informative and result in larger performance gains. Our emerging trend detection strategy is based on contrasting frequency of words in older data (training data) against frequency in newly collected data (recent data). More specifically, we formulate this task as detection of trending n-grams. We compute the trend scores for each n-gram, $n$, as follows:
\begin{equation*}
\left.\begin{aligned}
score(n) = \frac{f_{n, R} - f_{n, P}}{f_{n, P} + k}
\end{aligned}\right.
\end{equation*}
where $f_{n, R}$ and $f_{n, P}$ are the frequencies of n-gram $n$ in the recent and past datasets, respectively. In practical applications, $f_{n, R}$ can refer to the frequency in newly collected data, while $f_{n, P}$ can refer to the frequency in older data. $k$ is a normalization term used to mitigate the frequency of the highly-frequent n-grams in the recent datasets. When computing trend scores, we filter out stop words as they are usually the most common words but contain the least information. 
After we compute trend scores for all n-grams in newly collected data, we assign trend scores to the instances by summing over the scores of each n-gram in that instance (tweet). We then use the score to rank instances for labeling and updating the NER model. Our approach is flexible as it can be used in combination with any NER model architecture.

\input{figures/temporal_setting_100}
\input{figures/random_setting_100}

\section{Experiments}
\label{tab:experiments}
We empirically study the impact of retraining NER models on trending data in two different scenarios. In \textbf{Scenario 1}, we retrain the model in an incremental manner with $N$ instances from a newer batch of data in the following year at every iteration.  In \textbf{Scenario 2}, we retrain the model incrementally as well, but the pool of data we used to select instances includes all years available in the training partitions. In both cases, instances are selected based on their trend scores.  

We use the Temporal Twitter Dataset from \citet{rijhwani-preotiuc-pietro-2020-temporally} for all experiments. This dataset is temporally distributed and balanced with a variety of topics. It has 12K tweets collected from 2014 to 2019, with 2K samples from each year. In our experiments, the training set comprises of splits from 2014 to 2018. The validation set and test set have a random sample of 500 (25\%) and 1,500 (75\%) tweets from 2019, respectively.

\subsection{Neural Architectures}
\label{Neural Architectures}
As mentioned earlier, our approach is model agnostic. We validate this claim by experimenting with different NER neural architectures used in the prior art. The main difference between these models is the representation fed into a Conditional Random Field (CRF) \citep{Lafferty2001ConditionalRF} for prediction. The implementation and hyperparameters are described in Appendix \ref{Details for Experimental Setup}.

\paragraph{BiLSTM + CRF} 
Following \citet{ma-hovy-2016-end}, we use the GloVe \citep{pennington-etal-2014-glove} word embeddings for word representations and Convolution Neural Networks (CNNs) for character representations. Then a bidirectional LSTM \citep{Graves2005FramewisePC} takes both word representations and character representations as input and encodes sentences. 

\paragraph{BERT + CRF} 
BERT is a transformer-based model proposed by \citet{devlin-etal-2019-bert}. It is pre-trained using masked language modeling and next sentence prediction objectives on the corpora from the general domain. BERT takes subwords as input and generates contextualized word representations for each sentence. 

\paragraph{BERTweet + CRF} 
Similar to BERT, BERTweet \citep{nguyen-etal-2020-bertweet} is a large-scale language model with the same configuration as BERT. It is pre-trained on the corpora from the social media domain and achieves state-of-the-art results on many downstream Twitter NER tasks.

\subsection{Results}
\label{Results}
We empirically examine the performance of models under the influence of data evolution and temporal drift. We start with doing experiments on trending bi-grams and use the same amount of training samples at each step to eliminate the influence of training data size. Below we discuss the results of the two evaluation scenarios.

\paragraph{Scenario 1}
In this scenario, we assume that the data can only be accessed chronologically by year. For each new batch of data selected based on the trendiness score (\textbf{trend}), we take the model as trained on the previous batch and retrain on the newest data. In other words, we consider the model from the previous iteration as a pre-trained model and fine-tune that model on the newest data. For comparison purposes, we run a \textbf{temporal} version, where the model is fine-tuned with newer data every time, but the instances are selected randomly for the corresponding year. Due to the randomness in this approach, we run each model five times, and then we report the average of the five runs as the final F1 score.

The results are shown in Figure \ref{fig:temporal_setting_100}. We observe that both temporal and trend F1 scores increase as we move temporally closer to the target data. However, in all cases, the trend-based models always reach a higher score.

\paragraph{Scenario 2}
In this scenario, we assume the data can be accessed from all years at once. We merge the training data from all years and form a single pool of data. However, we still fine-tune models at each iteration using the same number of new instances each time. For the trend models, we select instances based on their trend scores, regardless of the year, whereas for the random model, we select instances at random from the merged pool of data. Similar to what we did in scenario 1, we run each model 5 times and report the averaged results.

The results are shown in Figure \ref{fig:random_setting_100}. Similar to scenario 1, the F1 scores of the models trained on instances selected based on their trend scores are always higher than random sampling F1 scores. In addition, scenario 2, on average, works better than scenario 1, which is consistent with \citet{rijhwani-preotiuc-pietro-2020-temporally}. However, this setting requires the data available from all years from the very beginning. Compared to scenario 2, scenario 1 is far more realistic because it can be more easily applied in practice.

\subsection{Analysis}

\paragraph{Impact of training data size}
We ran additional experiments where we add different amounts of training data at each iteration (50 and 200). With less training data available, the benefits of selecting instances based on trend scores are amplified. Even if more data is available, using trend scores to select which instances to add always results in better performance than randomly choosing instances. Due to space limitations, the plots are in Appendix \ref{Experiment with more data} figures \ref{fig:temporal_setting_50}, \ref{fig:random_setting_50}, \ref{fig:temporal_setting_200} and  \ref{fig:random_setting_200}. 

\paragraph{Impact of pre-trained knowledge}
From figures \ref{fig:temporal_setting_100} and \ref{fig:random_setting_100}, we observe that, in general, pre-trained models (BERT and BERTweet) tend to perform closer to that of the trend-based models. Apart from the well-documented advantages of contextualized representations, we believe that higher performance here is due to these models' pre-trained knowledge. We suspect that if we had the ability to control the data, and in particular, the year of the data used in pre-training, the results would be different, and we would observe a larger gap between pre-trained transformer models and the trend-based approach.

\paragraph{Entity-wise Model Performance}
We investigate whether our approach affects named entity types differently. To this end, we create random data and trending data. The random data is randomly selected, while the trending data is selected based on the trend scores. Each data has 1,000 samples. Table \ref{tab: performance_comparison} shows the model performance on the random data, versus the trending data. We notice that all three models overall benefit from trend detection with an improvement from 2.70\% ~ 5.71\% on F1 metric, indicating that the models can adequately learn the context of named entities.

\input{tables/performance_comparison}

To better understand the high model performance on trending data, Table \ref{tab:data_distribution} shows the distribution of random and trending data. By selecting training samples based on our approach, the number of entities in the trending data is 77\% more than the number of entities in the random data, including 92\% more PER, 38\% more LOC, and 91\% more ORG. In the token level, there are more 108\% entity tokens in the trending data than in the random data. The higher ratio of named entities in the trending data increases the diversity of each entity type, and therefore, decreases the test error. 

\input{tables/data_distribution}

\section{Related Work}
Previous work has studied trend detection in online social media platforms such as Twitter and Facebook \citep{Benhardus2013StreamingTD, Mathioudakis2010TwitterMonitorTD, Miot2020AnES}. \citet{Benhardus2013StreamingTD} outlined the methodologies for using the data from online platforms and proposed criteria based on the frequency of words to identify trending topics in Twitter. \citet{Mathioudakis2010TwitterMonitorTD} presented a system to detect bursty keywords that suddenly appear in tweets at an unusually high rate. Recently, \citet{Miot2020AnES} investigated the efficiency of deep neural networks to detect trends. However, these techniques are applied without taking named entities into consideration.

Towards emerging named entities, recent work has mainly focus on identification and classification of unusual and previously unseen named entities. \citet{derczynski-etal-2015-usfd} investigated the effects of data drift and the evaluation of the NER models on temporally unseen data. \citet{agarwal-etal-2018-dianed} studied the disambiguation of named entities with explicit consideration of temporal background. \citet{rijhwani-preotiuc-pietro-2020-temporally} reported improvements on performance for overlapping named entities under the impact of temporal drift. Due to the limitation of resources and lack of annotated data from social media, these NER models tend to have lower performances on emerging named entities.

\section{Conclusion}
In this work, we propose a simple approach to update model parameters and prevent degradation performance from temporal drifts. Our approach is inspired by our observations of how Twitter data follows trends in topics that can change very quickly. Experimentally, we show that leveraging emerging trends can benefit the recognition of named entities and reduce performance degradation, especially in low-resource scenarios. Our proposal is model agnostic, and can potentially be adapted to other NLP tasks that target social media and face the same problems of data evolution and temporal drift.

\section*{Acknowledgements}
This work was partially supported by the National Science Foundation (NSF) under grant \#1910192. We would like to thank the members from the RiTUAL lab at the University of Houston for their invaluable feedback. We also thank the anonymous reviewers for their valuable suggestions.

\bibliography{anthology,custom}
\bibliographystyle{acl_natbib}

\appendix

\input{figures/temporal_setting_50}
\input{figures/random_setting_50}
\input{figures/temporal_setting_200}
\input{figures/random_setting_200}

\section{Details for Experimental Setup}
\label{Details for Experimental Setup}
For BiLSTM-CRF model, we use GloVe Twitter embeddings. The dimensions of character embeddings and word embeddings are 50 and 100 respectively. We then use 2-layer LSTM with 300 hidden units to encode sentences. The dropout rate is 0.5. During training, we use stochastic gradient descent (SGD) with learning rate 0.1, batch size 20, and momentum 0.9. The L2 regularization is set to 0.001. For BERT and BERTweet, we do fine-tuning using AdamW optimizer \citep{loshchilov2017decoupled} with learning rate 5e-5, batch size 32, and weight decay 0.01. We also use a gradient clipping of 1.0 and the dropout rate is 0.1. In scoring function, $k$ is set as 0.1 for sample selection.

\section{Experiment with more data}
\label{Experiment with more data}
In Figure \ref{fig:temporal_setting_50} and Figure \ref{fig:random_setting_50}, we use 50 instances at each step. In Figure \ref{fig:temporal_setting_200} and Figure \ref{fig:random_setting_200}, we use 200 instances at each step. We repeat our experiment with using different number of instances at each training step to study the impact of dataset size.

\end{document}

%% file: figures/example_of_tweets.tex
\begin{figure}[t]
\centering
\subfigure{
    \begin{minipage}[t]{1.0\linewidth}
    \centering
    \includegraphics[width=1\linewidth]{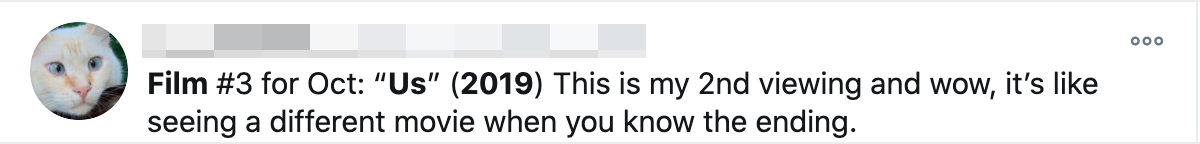}
    \end{minipage}
}
\subfigure{
    \begin{minipage}[t]{1.0\linewidth}
    \centering
    \includegraphics[width=1\linewidth]{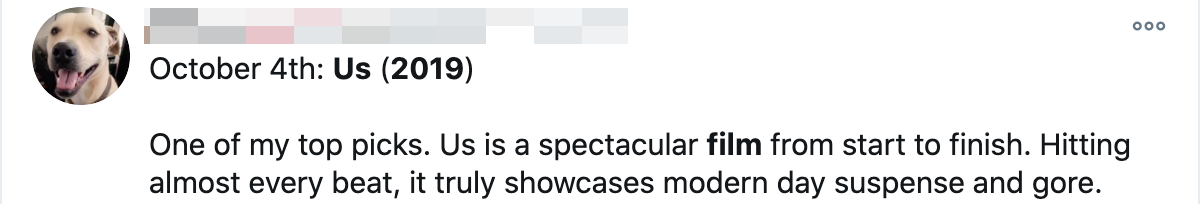}
    \end{minipage}
}
\subfigure{
    \begin{minipage}[t]{1.0\linewidth}
    \centering
    \includegraphics[width=1\linewidth]{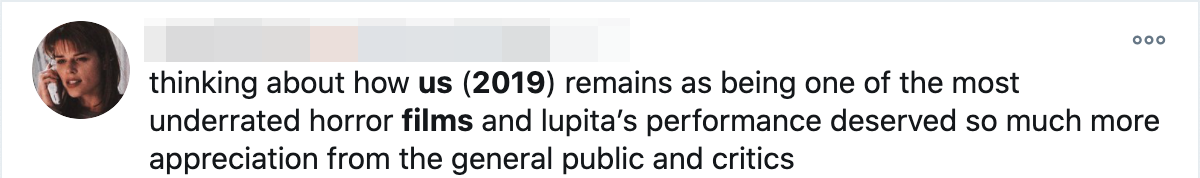}
    \end{minipage}
}
\caption{Examples of tweets that include the emerging topic ‘\textbf{US}’, a horror movie released in 2019}
\label{fig:example_of_tweets}
\end{figure}

%% file: figures/temporal_setting_100.tex
\begin{figure*}[ht]
\centering
\subfigure[BiLSTM + CRF]{
    \begin{minipage}[t]{0.31\linewidth}
    \centering
    \includegraphics[width=1\linewidth]{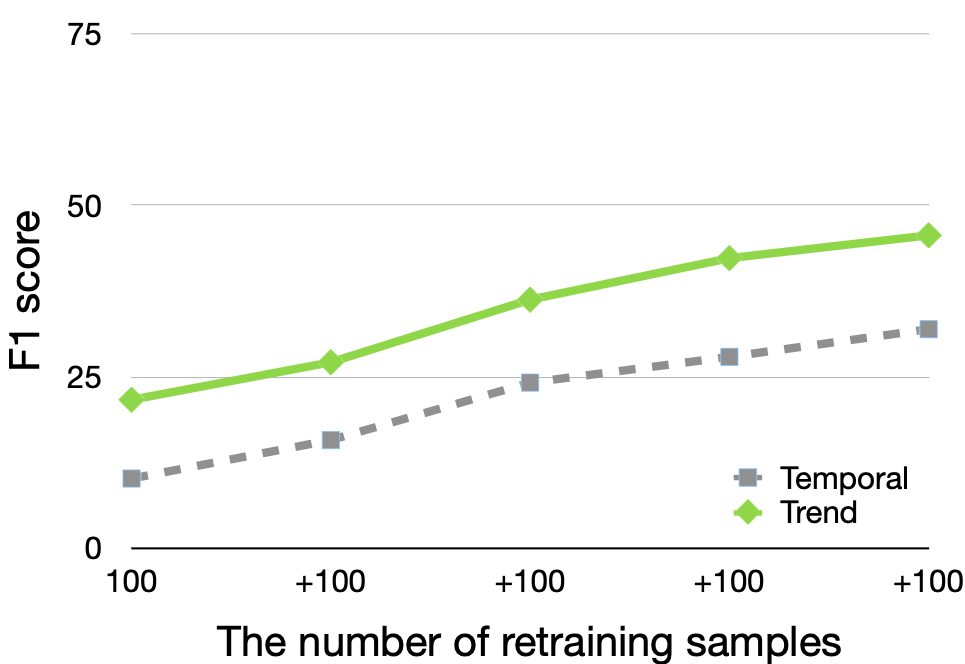}
    \end{minipage}
}
\subfigure[BERT + CRF]{
    \begin{minipage}[t]{0.31\linewidth}
    \centering
    \includegraphics[width=1\linewidth]{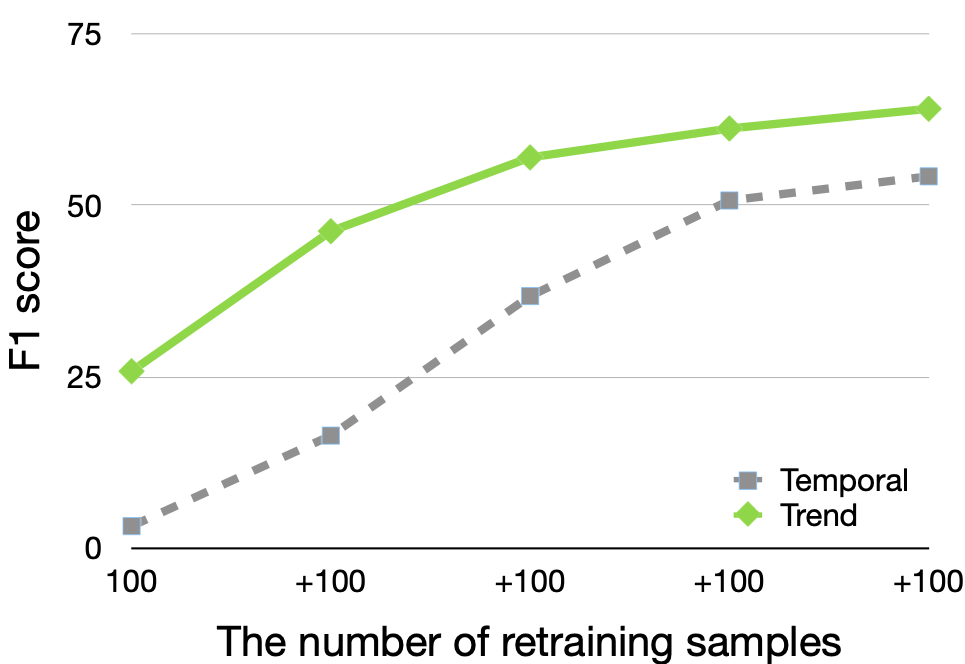}
    \end{minipage}
}
\subfigure[BERTweet + CRF]{
    \begin{minipage}[t]{0.31\linewidth}
    \centering
    \includegraphics[width=1\linewidth]{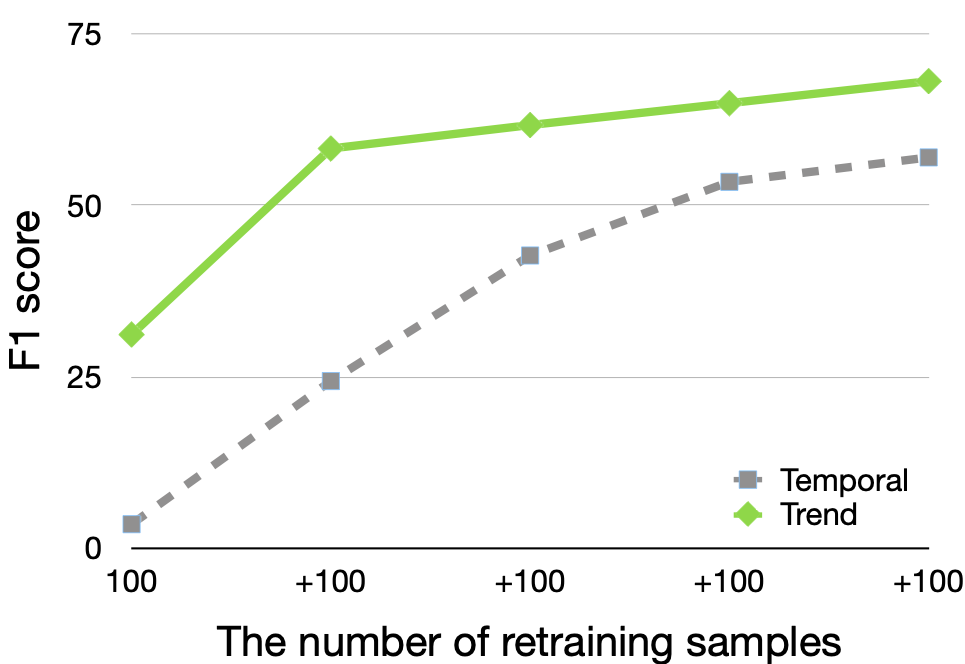}
    \end{minipage}
}
\caption{\textbf{Data can only be accessed year by year - } Each step represents a year from 2014 to 2018. At each step, we add instances from its respective year to the training set. For Temporal, we randomly select instances from that given year. For Trend, we rank instances based on their trending score. We experiment with 50 (Appendix \ref{Experiment with more data}), 100 and 200 (Appendix \ref{Experiment with more data}) instances per step to show the impact of training size.}
\label{fig:temporal_setting_100}
\end{figure*}

%% file: figures/random_setting_100.tex
\begin{figure*}[ht]
\centering
\subfigure[BiLSTM + CRF]{
    \begin{minipage}[t]{0.31\linewidth}
    \centering
    \includegraphics[width=1\linewidth]{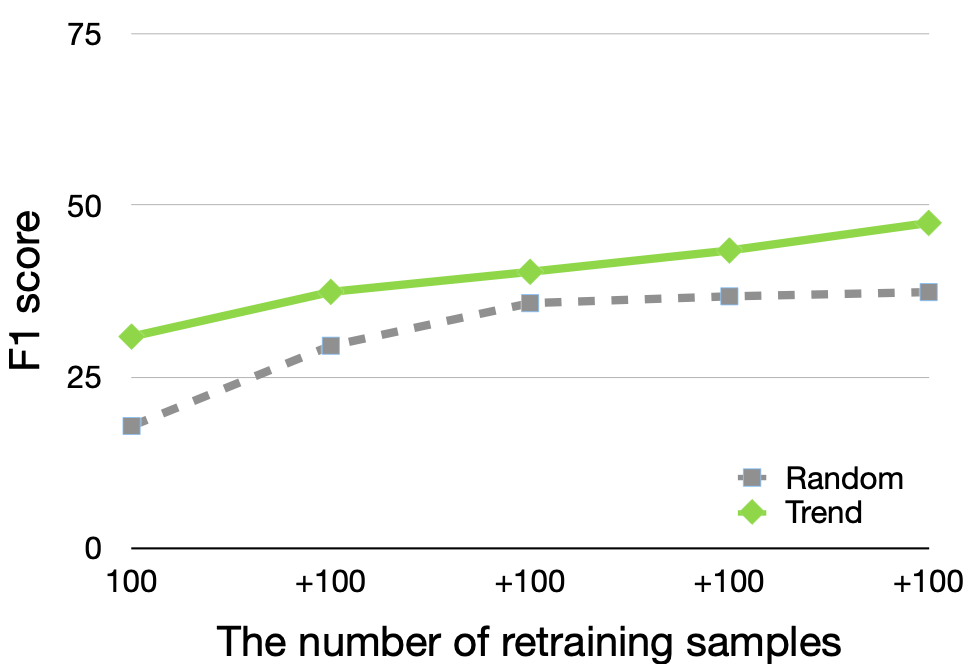}
    \end{minipage}
}
\subfigure[BERT + CRF]{
    \begin{minipage}[t]{0.31\linewidth}
    \centering
    \includegraphics[width=1\linewidth]{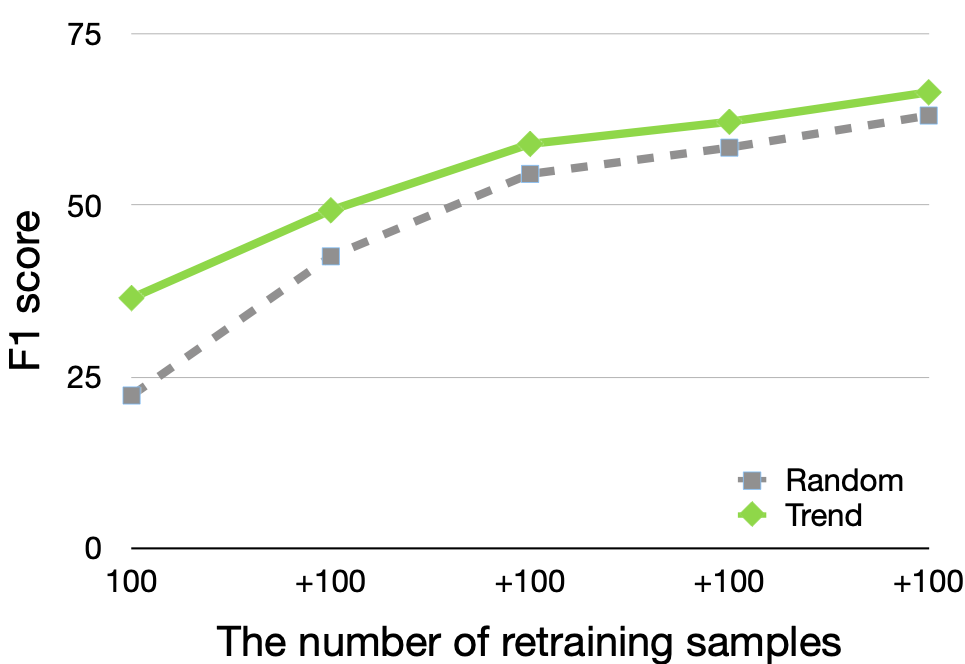}
    \end{minipage}
}
\subfigure[BERTweet + CRF]{
    \begin{minipage}[t]{0.31\linewidth}
    \centering
    \includegraphics[width=1\linewidth]{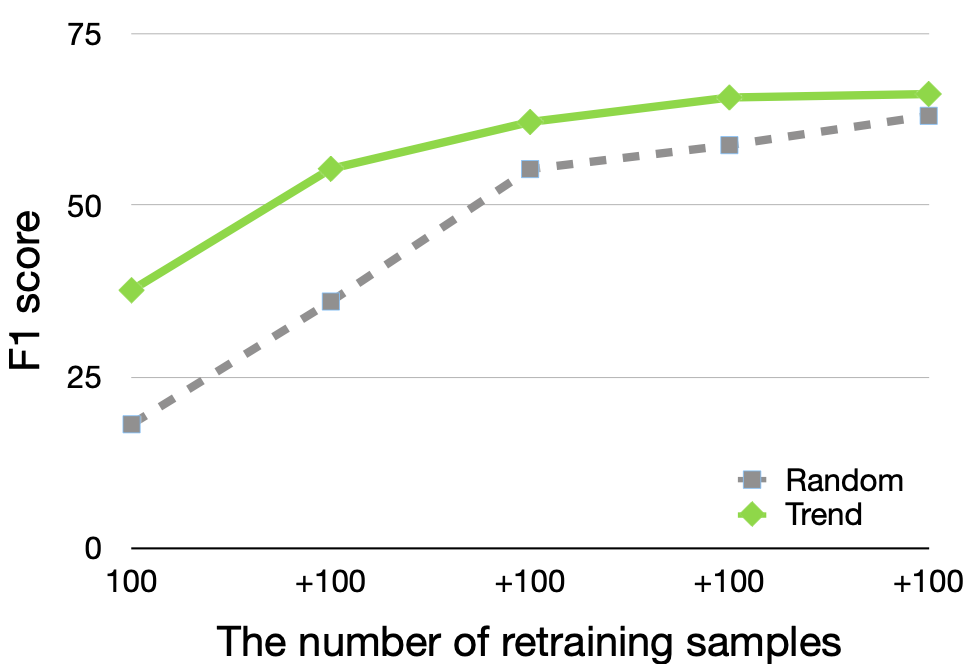}
    \end{minipage}
}
\caption{\textbf{Data from all years is available -} At each step, we add new instances to our training set. For Random, we randomly select instances from the available data. For Trend, we rank all available instances from most trending to less trending based on their trending scores. We then use this ranking to select the instances. At each step, we choose the instances with the highest trending scores that have not yet been added to the training set. We experiment with 50 (Appendix \ref{Experiment with more data}), 100 and 200 (Appendix \ref{Experiment with more data}) instances per step to show the impact of training size.}
\label{fig:random_setting_100}
\end{figure*}

%% file: tables/performance_comparison.tex
\begin{table}[ht]
    \small
    \centering
    \renewcommand{\arraystretch}{1.3}
    \resizebox{0.45\textwidth}{!}{
    \begin{tabular}{lcccccc}
        \toprule
        \multirow{2}{*}{\bf Model}  & \multicolumn{3}{c}{\bf Random data}       & \multicolumn{3}{c}{\bf Trending data}\\ 
        \cmidrule(lr){2-4} \cmidrule(lr){5-7}
        & \bf P & \bf R & \bf F1 & \bf P & \bf R & \bf F1 \\
        \toprule
        BiLSTM + CRF                & 59.38 & 48.02 & 53.10 & 63.42 & 54.83 & 58.81 \\
        BERT + CRF                  & 62.26 & 73.23 & 67.30 & 70.07 & 69.93 & 70.00 \\
        BERTweet + CRF              & 60.64 & 64.84 & 62.67 & 65.45 & 70.46 & 67.86 \\
        \toprule
        \end{tabular}
    }
    \caption{Performance comparison on random data and trending data, including persc.}
    \label{tab: performance_comparison}
\end{table}

%% file: tables/data_distribution.tex
\begin{table}[ht]
    \centering
    \small
    \renewcommand{\arraystretch}{1.3}
    \resizebox{0.45\textwidth}{!}{
    \begin{tabular}{lcccc}
        \toprule
        \multirow{2}{*}{\bf Entity Type}& \multicolumn{2}{c}{\bf Random data}   & \multicolumn{2}{c}{\bf Trending data} \\ \cmidrule(lr){2-3} \cmidrule(lr){4-5}
                                        & \bf Entity-level & \bf Token-level & \bf Entity-level & \bf Token-level\\
        \toprule
        PER & 225 & 340 & 432 & 755\\
        LOC & 178 & 226 & 245 & 362\\
        ORG & 281 & 379 & 537 & 848\\
        \toprule
        Total & 684 & 945 & 1,214 & 1,965 \\
        \toprule
    \end{tabular}
    }
    \caption{The distribution of random data and trending data, including entity-level distribution (entity spans) and token-level distribution (entity tokens).}
    \label{tab:data_distribution}
\end{table}

%% file: figures/temporal_setting_50.tex
\begin{figure*}[ht]
\centering
\subfigure[BiLSTM + CRF]{
    \begin{minipage}[t]{0.31\linewidth}
    \centering
    \includegraphics[width=1\linewidth]{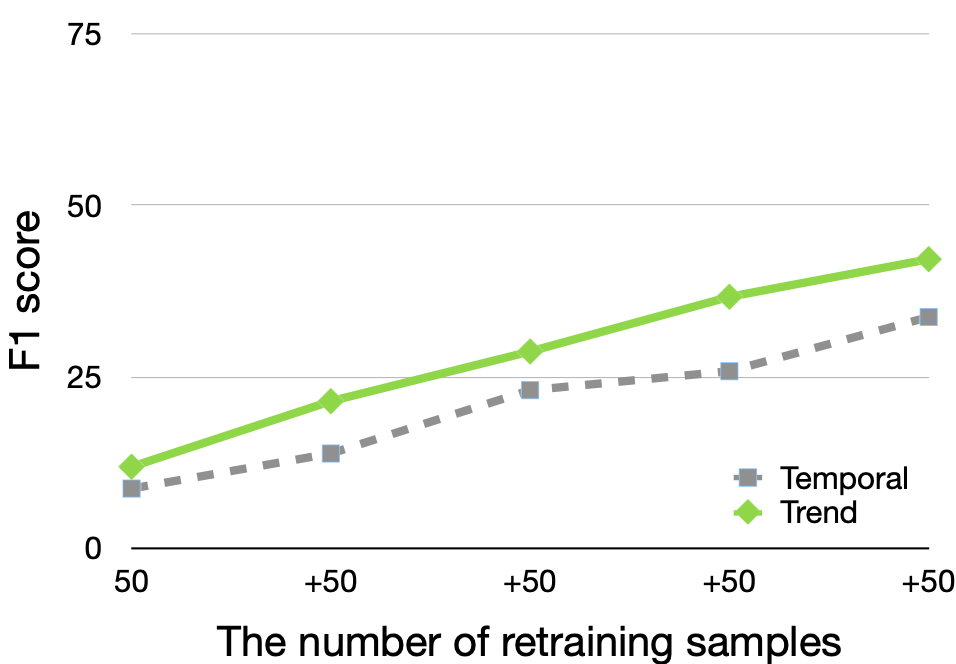}
    \end{minipage}
}
\subfigure[BERT + CRF]{
    \begin{minipage}[t]{0.31\linewidth}
    \centering
    \includegraphics[width=1\linewidth]{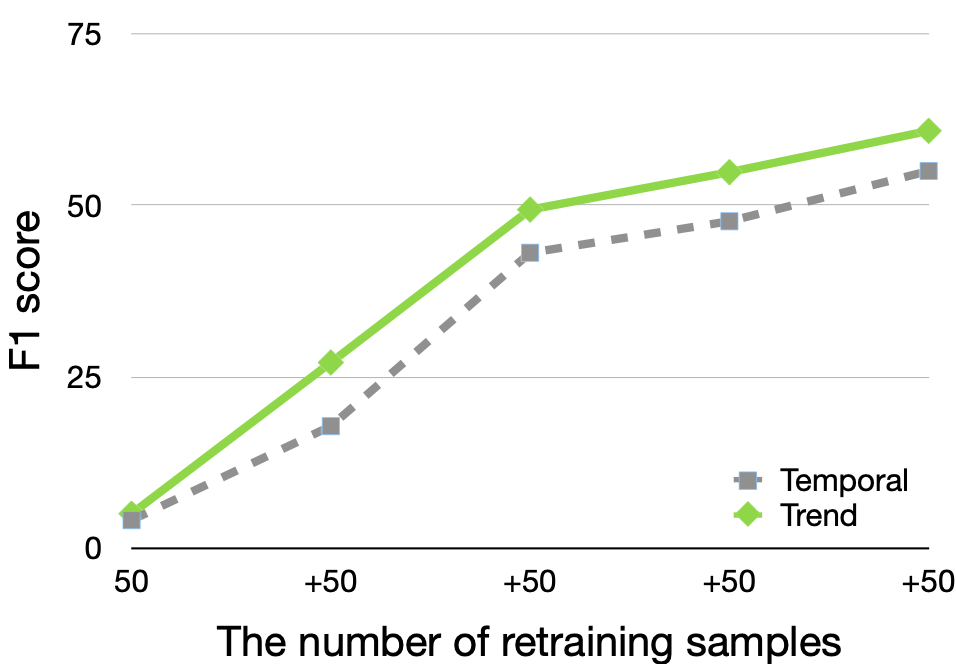}
    \end{minipage}
}
\subfigure[BERTweet + CRF]{
    \begin{minipage}[t]{0.31\linewidth}
    \centering
    \includegraphics[width=1\linewidth]{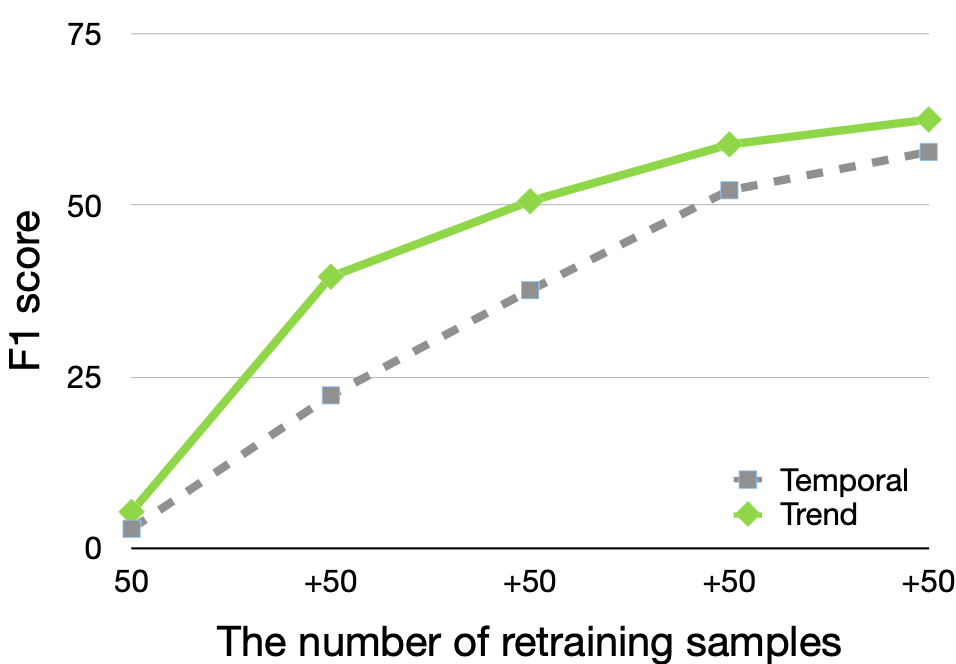}
    \end{minipage}
}
\vspace{-0.5cm}
\caption{\textbf{Data can only be accessed year by year - } Each step represents a year from 2014 to 2018. At each step, we add instances from its respective year to the training set. For Temporal, we randomly select instances from that given year. For Trend, we rank instances based on their trending score. We experiment with 50 (Appendix \ref{Experiment with more data}), 100 and 200 (Appendix \ref{Experiment with more data}) instances per step to show the impact of training size.}
\label{fig:temporal_setting_50}
\vspace{-0.5cm}
\end{figure*}

%% file: figures/random_setting_50.tex
\begin{figure*}[ht]
\centering
\subfigure[BiLSTM + CRF]{
    \begin{minipage}[t]{0.31\linewidth}
    \centering
    \includegraphics[width=1\linewidth]{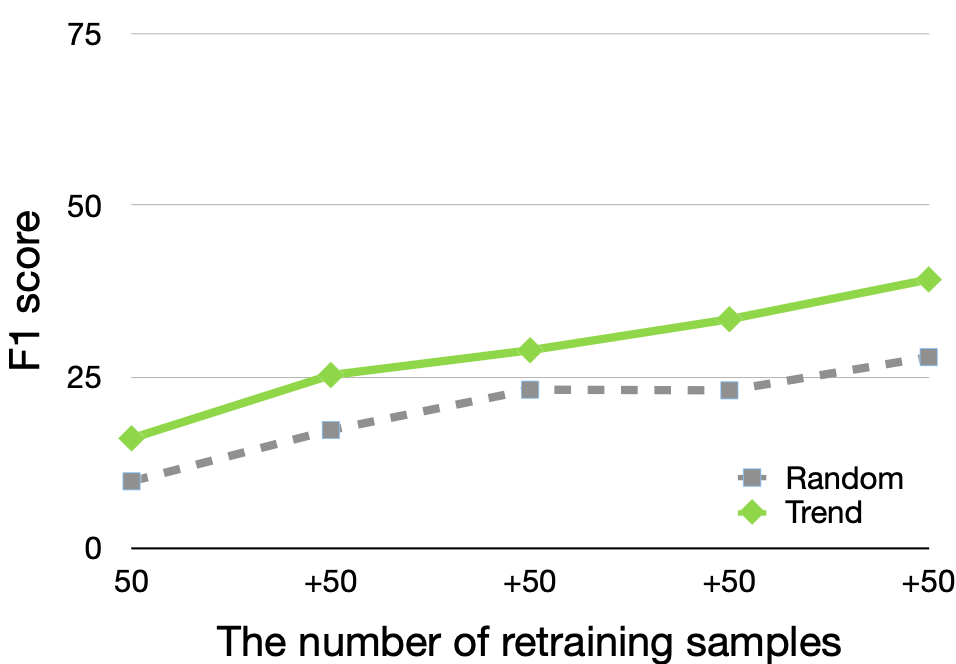}
    \end{minipage}
}
\subfigure[BERT + CRF]{
    \begin{minipage}[t]{0.31\linewidth}
    \centering
    \includegraphics[width=1\linewidth]{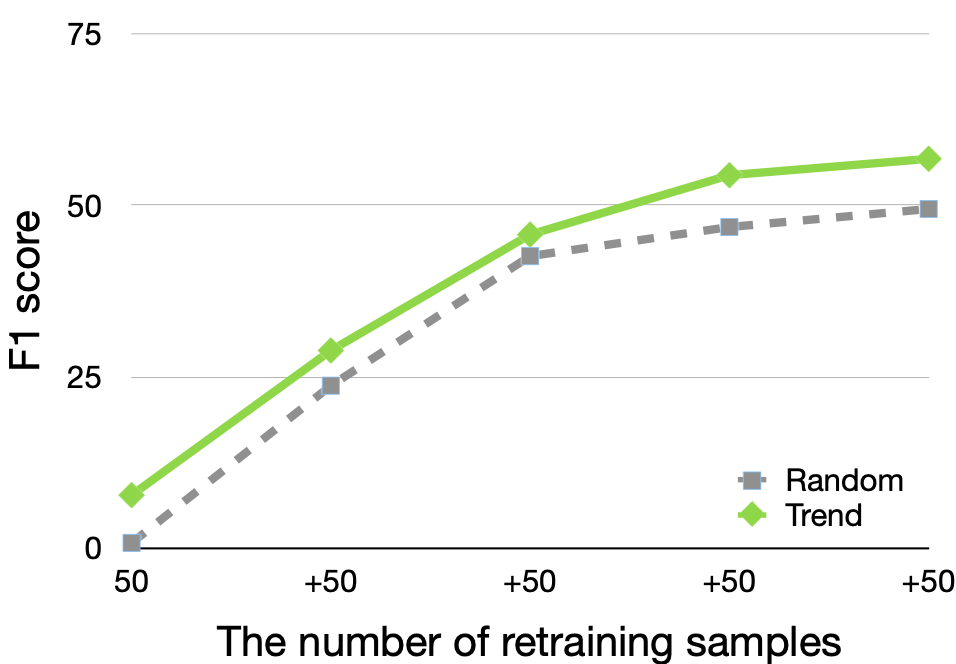}
    \end{minipage}
}
\subfigure[BERTweet + CRF]{
    \begin{minipage}[t]{0.31\linewidth}
    \centering
    \includegraphics[width=1\linewidth]{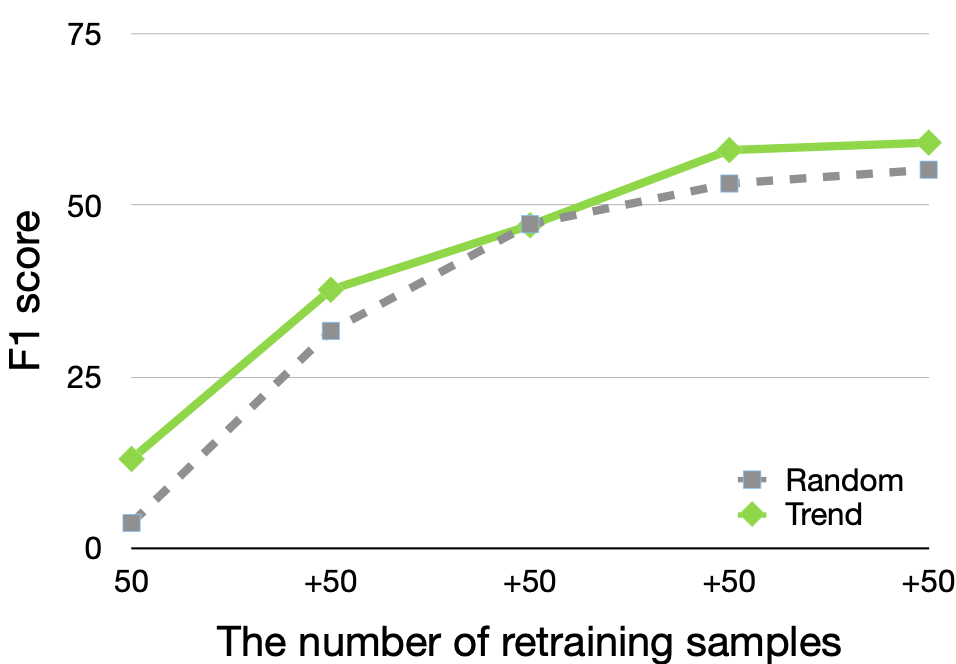}
    \end{minipage}
}
\vspace{-0.5cm}
\caption{\textbf{Data from all years is available -} At each step, we add new instances to our training set. For Random, we randomly select instances from the available data. For Trend, we rank all available instances from most trending to less trending based on their trending scores. We then use this ranking to select the instances. At each step, we choose the instances with the highest trending scores that have not yet been added to the training set. We experiment with 50 (Appendix \ref{Experiment with more data}), 100 and 200 (Appendix \ref{Experiment with more data}) instances per step to show the impact of training size.}
\label{fig:random_setting_50}
\vspace{-0.5cm}
\end{figure*}

%% file: figures/temporal_setting_200.tex
\begin{figure*}[t!]
\centering
\subfigure[BiLSTM + CRF]{
    \begin{minipage}[t]{0.31\linewidth}
    \centering
    \includegraphics[width=1\linewidth]{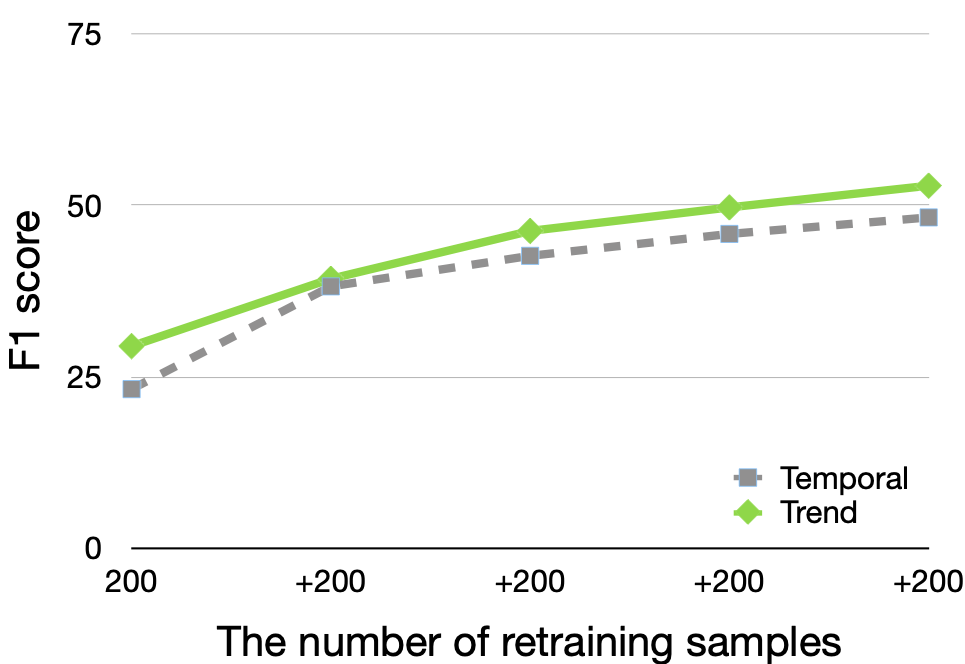}
    \end{minipage}
}
\subfigure[BERT + CRF]{
    \begin{minipage}[t]{0.31\linewidth}
    \centering
    \includegraphics[width=1\linewidth]{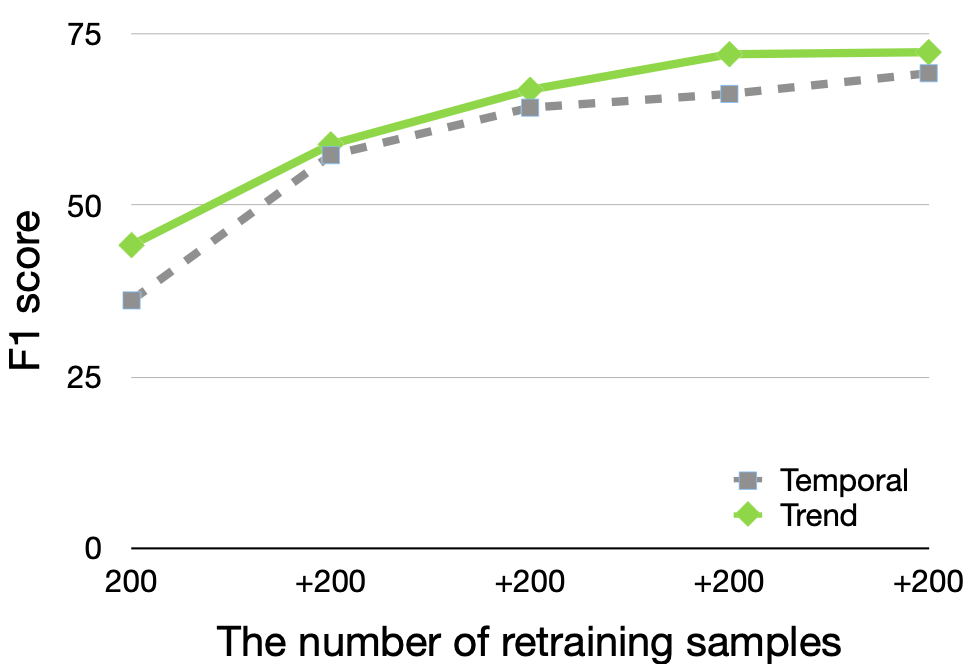}
    \end{minipage}
}
\subfigure[BERTweet + CRF]{
    \begin{minipage}[t]{0.31\linewidth}
    \centering
    \includegraphics[width=1\linewidth]{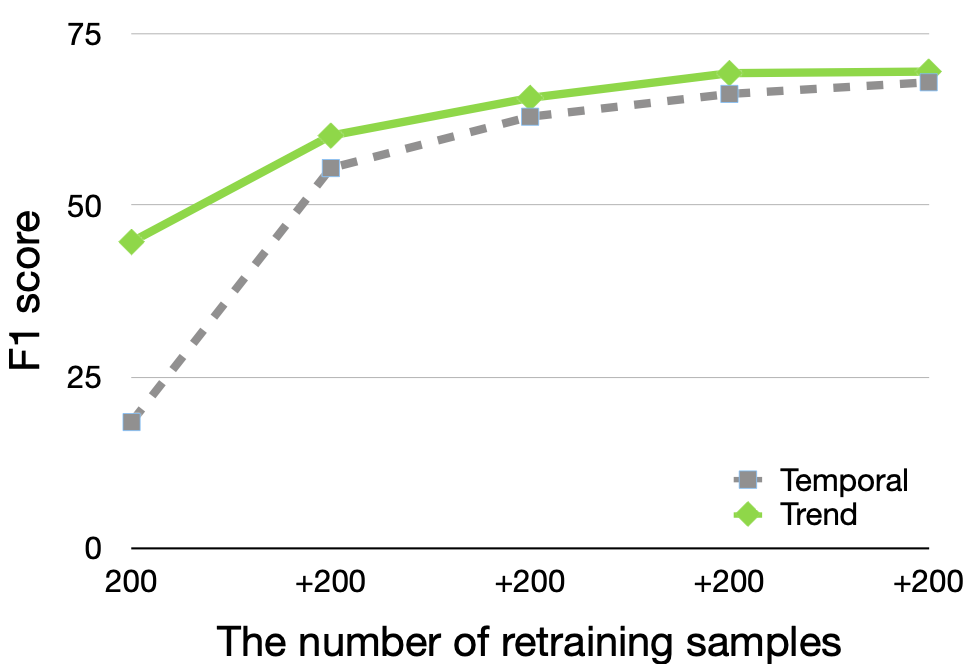}
    \end{minipage}
}
\vspace{-0.5cm}
\caption{\textbf{Data can only be accessed year by year - } Each step represents a year from 2014 to 2018. At each step, we add instances from its respective year to the training set. For Temporal, we randomly select instances from that given year. For Trend, we rank instances based on their trending score. We experiment with 50 (Appendix \ref{Experiment with more data}), 100 and 200 (Appendix \ref{Experiment with more data}) instances per step to show the impact of training size.}
\label{fig:temporal_setting_200}
\vspace{-0.5cm}
\end{figure*}

%% file: figures/random_setting_200.tex
\begin{figure*}[t!]
\centering
\subfigure[BiLSTM + CRF]{
    \begin{minipage}[t]{0.31\linewidth}
    \centering
    \includegraphics[width=1\linewidth]{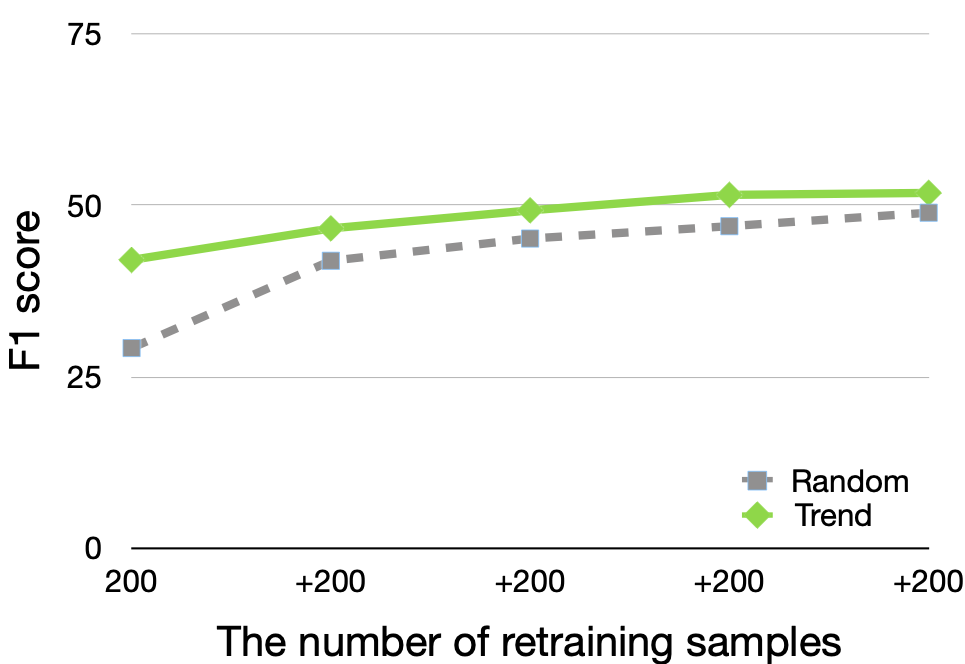}
    \end{minipage}
}
\subfigure[BERT + CRF]{
    \begin{minipage}[t]{0.31\linewidth}
    \centering
    \includegraphics[width=1\linewidth]{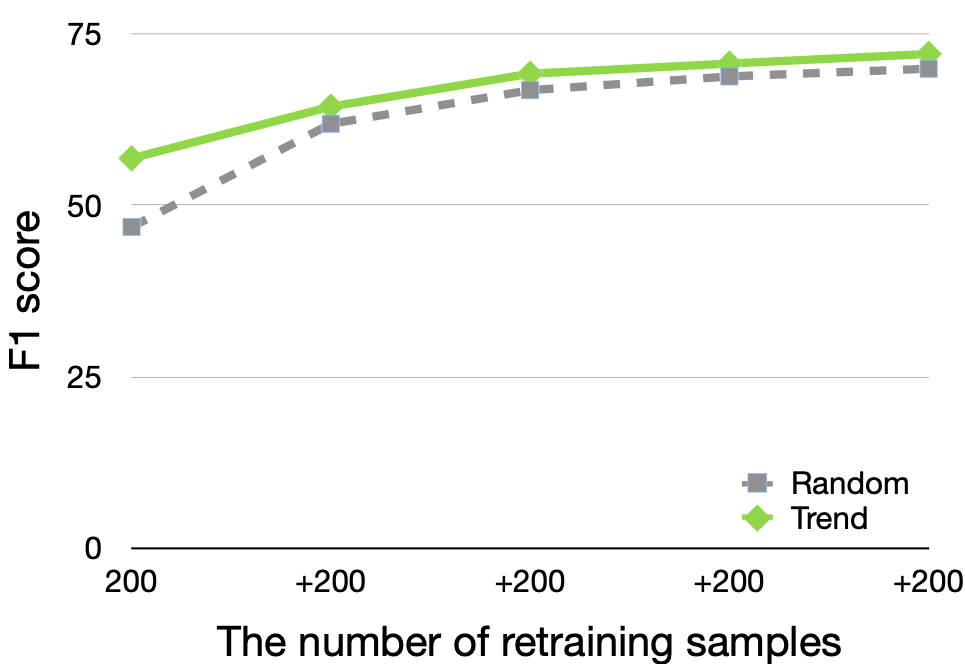}
    \end{minipage}
}
\subfigure[BERTweet + CRF]{
    \begin{minipage}[t]{0.31\linewidth}
    \centering
    \includegraphics[width=1\linewidth]{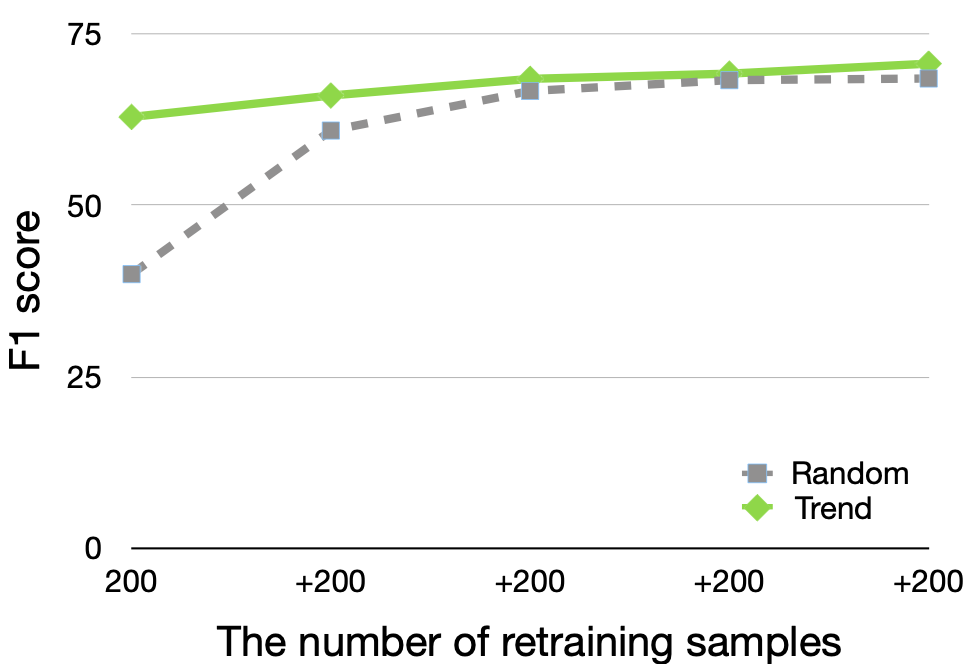}
    \end{minipage}
}
\vspace{-0.5cm}
\caption{\textbf{Data from all years is available -} At each step, we add new instances to our training set. For Random, we randomly select instances from the available data. For Trend, we rank all available instances from most trending to less trending based on their trending scores. We then use this ranking to select the instances. At each step, we choose the instances with the highest trending scores that have not yet been added to the training set. We experiment with 50 (Appendix \ref{Experiment with more data}), 100 and 200 (Appendix \ref{Experiment with more data}) instances per step to show the impact of training size. }
\label{fig:random_setting_200}
\vspace{-0.5cm}
\end{figure*}